\documentclass{article}

\usepackage[final]{nips_2018}

\usepackage[utf8]{inputenc} 
\usepackage[T1]{fontenc}    
\usepackage{hyperref}       
\usepackage{url}            
\usepackage{booktabs}       
\usepackage{amsfonts}       
\usepackage{nicefrac}       
\usepackage{microtype}      
\usepackage{graphicx,epstopdf}
\usepackage[usenames, dvipsnames]{color}

\title{Generating User-friendly Explanations for Loan Denials using GANs}

\author{
  \And
  Ramya Srinivasan  \thanks{The authors contributed equally to the work} \\
  Fujitsu Laboratories of America\\
  \texttt{ramya@us.fujitsu.com} \\
  \And
  Ajay Chander $^*$ \\
  Fujitsu Laboratories of America\\
  \texttt{ajayc@us.fujitsu.com} \\
  \AND
    Pouya Pezeshkpour $^*$
  \\
  University of California, Irvine\\
  \texttt{pezeshkp@uci.edu} \\
  }

\begin{document}

\maketitle

\begin{abstract}

Financial decisions impact our lives, and thus everyone from the regulator to the consumer is interested in fair, sound, and explainable decisions. There is increasing competitive desire and regulatory incentive to deploy AI mindfully within financial services.  An important mechanism towards that end is to explain AI decisions to various stakeholders. State-of-the-art explainable AI systems mostly serve AI engineers and offer little to no value to business decision makers, customers, and other stakeholders. Towards addressing this gap, in this work we consider the scenario of explaining loan denials.  We build the first-of-its-kind dataset that is representative of loan-applicant friendly explanations. We design a novel Generative Adversarial Network (GAN) that can accommodate smaller datasets, to generate user-friendly textual explanations. We demonstrate how our system can also generate explanations serving different purposes: those that help {\it educate} the loan applicants, or help them take appropriate {\it action} towards a future approval. 
We hope that our contributions will aid the deployment of AI in financial services by serving the needs of the wider community of users seeking explanations.


\end{abstract}

\section{Introduction}

From customer behavior prediction and identity verification to fraud detection and intelligent chatbots that can answer customer queries, AI is being deployed within a wide variety of Fintech applications. The total funding for AI-based Fintech projects worldwide reached USD 8.2 billion in the third quarter of 2017 \cite{news}. With massive funding and extensive experimentation, it's crucial for financial companies to use AI mindfully within their services to stay competitive.

Amidst this wide-spread business adoption of AI, customers, policymakers and technologists are getting increasingly concerned about AI being a blackbox technology. For relatively benign high volume AI decisions, such as those in online recommendation systems, an opaque yet accurate algorithm provides commercial value with few immediate downsides. However in more critical domains such as healthcare, finance, and law and order, a blackbox AI is unacceptable.

Consider, for example, an AI-based credit scoring system. In markets where credit risk scoring models are regulated and scrutinized, there is a strong requirement for the models and the credit decisions derived from them, to be explainable. The impact each variable has on the credit score must be clearly explained to, and be acceptable to lenders, regulators, and consumers \cite{fico}. 

As a result, there have been several initiatives both from the government   \cite{darpa} and a diverse set of industries \cite{kyndi,pwc} to make AI explainable. Yet, most state-of-the-art methods provide explanations that mostly target the needs of AI engineers \cite{selvaraju,park}. In other words, explanations assume some domain knowledge or are generated for people with domain expertise. As the use of AI becomes widespread, there is an increasing need for creating AI systems that can explain their decisions to a large community of users who are not necessarily domain experts. These users could include software engineers, business owners, and end-users. 

As an instance, consider the use case of AI-based loan decisions. From bank managers, investors, and loan applicants, to credit bureaus and regulatory agencies, a variety of stakeholders participate in or have interest in this financial service. The nature of explanation sought differs based on the role of the human-in-the-loop requiring the explanation. For example, while an explanation like ``Person A was denied this loan as their credit score was similar to Person B" might be helpful to an AI engineer, it offers little to no value to a loan applicant. By considering the overall context in which an explanation is presented, including the role played by the human-in-the-loop, we can hope to craft effective explanations.

Effective explanations leverage the cognitive and social processes by which people create meaning, build trust, and make decisions. In addition to  addressing issues such as regulation, and aiding in adopting good practices around accountability and ethics, effective explanations serve a variety of purposes. These include building {\it trust} – using explainable AI (XAI) systems provides greater visibility over unknown vulnerabilities and can assure stakeholders that the system is operating as desired. Explanations can also aid in understanding why and how a model works the way it does and help in {\it debugging} and {\it designing} it for better robustness. Explanations can also help in {\it educating} about business drivers such as revenue, cost, customer behaviour, and employee turnover, to improve strategy and to take appropriate {\it actions}. 

The focus of this work is on crafting effective explanations that are suitable for end-users. We consider the use case of AI-based loan decisions, in particular, loan denials. We study what sort of explanations are acceptable to loan applicants and design a machine learning system that can generate end-user friendly explanations. Further, the system can generate explanations serving different purposes such as those that help in {\it educating} the loan applicant, and those that help the applicant in taking appropriate {\it action} towards a future approval. The contributions are elaborated below.

\subsection{Contributions}
Our goal is to help bridge the gap between research and practice in the financial industry by serving the needs of the wider community of users seeking explanations and by generating explanations that serve different purposes (e.g. in educating, help in taking an action, etc.). There are at least three important contributions associated with this work: 

{\it 1) We collect and build a (first-of-its-kind) dataset that is representative of user-friendly explanations:}

Given that existing explainable AI methods mostly offer value to AI engineers, there is a need to understand the kind of explanations sought by end-users (e.g., loan applicants). Thus, we ran a survey on Amazon Mechanical Turk (AMT) and gathered the first-of-its-kind dataset to understand the nature of acceptable explanations for loan applicants. Surprisingly, we observed that most of the reasons provided by end-users seldom appeared as features in datasets used to train machine learning models. This, in turn, prompted us to explore this explainability problem using data provided by end-users. 

{\it 2) We design a novel GAN architecture to generate explanations using very limited training data: }

Next, we designed a novel conditional generative adversarial network (GAN) to generate user-friendly textual explanations based on the reasons mentioned in the collected dataset, which consisted of about 2432 sentences. Unlike existing GAN based text generation architectures that require large datasets for training, the proposed model is capable of generating text from as little as a thousand sentences, a good three orders of magnitude less than existing architectures.  We address the challenge of limited training data by means of three architectural modifications to \cite{zhao2017adversarially} listed below:

a) We incorporate a Gaussian mixture model for the noise input of the generator as proposed in \cite{gurumurthy2017deligan}. \\
 b) We embed multiple conditional information in a hierarchical manner.\\
 c) Motivated by the idea proposed in \cite{kocaoglu2017causalgan}, we introduce two new loss functions to classify sentences based on the reasons for loan denials.

{\it 3) We generate explanations serving different purposes: to educate and to help take action:}

As described earlier, based on the needs of the task and the human-in-the-loop seeking the explanation, an explanation has different functions: to educate, to help build trust, to design, to help take action, etc. The proposed model is capable of generating explanations for different purposes: {\it to educate loan applicants and to help them in taking appropriate actions}. 

It is to be noted that the proposed system can be combined with existing explainable AI systems so as to render a user-friendly explanation given an AI-engineer friendly explanation such as \cite{lime}. However, as mentioned earlier, the reasons provided by end-users seldom appeared as features in datasets that were used to train models. For this reason, we did not consider the task of building this pipeline and instead focused on the challenging problem of generating user-friendly explanations. 

The rest of the paper is organized as follows: Section 2 provides a review of related work. The dataset is described in Section 3. Details of the proposed system are provided in Section 4.  Section 5 provides an account of the experiments and summarizes the results. We offer our conclusions in Section 6.

\section{Related Work}
We review related efforts, with emphasis on the finance industry. We then review related works in explainability from AI engineer and end-user perspectives.

\subsection{XAI Efforts in Financial Industry} The new European General Data Protection Regulation (GDPR and ISO/IEC 27001) and the U.S. Defense Advanced Research Projects Agency's XAI program \cite{darpa} are some note-worthy governmental initiatives towards explainable AI. In parallel, several industry groups are looking to address issues concerning AI explainability. Within the financial sector, leading institutions such as Capital One, JC Morgan and FICO are investing heavily in AI. For example, Adam Wenchel, vice president of machine learning and data innovation at Capital One, said that the company would like to use deep learning for all sorts of functions, including deciding who is granted a credit card. Capital One also created a research team, led by Wenchel, dedicated to finding ways of making AI more explainable. FICO recently released an XAI toolkit \cite{xaitool} to outline some of the explainability support for their machine learning. The momentum for explainable AI is only expected to grow in the times to come.

\subsection{Explanations for the AI Engineer} A nice summary concerning explainability from an AI engineer's perspective is provided in \cite{lipton} and \cite{doran}. In \cite{selvaraju}, the authors highlight the regions in an image that were most important to the model in classifying the image. However, such explanations are not useful to an end-user in either understanding the AI's decision or in debugging the model \cite{chandrashekaran}.  In \cite{velez}, the authors discuss the main factors used by the AI system in arriving at a certain decision and also discuss how changing a factor changes the decision.  This kind of explanation helps in debugging for the AI engineers.  While impressive in helping an AI engineer, these works are not accessible to a wider set of users.

\subsection{Explanations for the End-User} More recently, there have been efforts in understanding the human interpretability of AI systems. The authors in \cite{finale} provide a taxonomy for human interpretability of AI systems.  A non-AI engineer perspective regarding explanations of AI system is provided in \cite {miller}. A nice perspective of user-centered explanations is provided in \cite{herman}, wherein the author emphasizes the need for persuasive explanations. The authors in \cite{amershi} explore the notion of interactivity from the lens of the user. In \cite{narayanan}, the authors discuss how humans understand explanations from machine learning systems through a user study. The metrics used to measure human interpretability are those concerning explanation length, number of concepts used in the explanation, and the number of repeated terms. Interpretability is measured in terms of the time to response and the accuracy of the response. While these efforts are significant in quantifying human interpretability, they fall short of generating user-friendly explanations, which is the focus of this work. 
\section{Dataset}
Due to the non-availability of any dataset that is representative of user-friendly explanations, we collected a first-of-its-kind dataset in this regard. We refer the dataset as ``{\bf X-Net}''. 

In order to build X-Net, we ran a survey on Amazon Mechanical Turk (AMT) wherein we provided the AMT workers with a loan application scenario and asked them to imagine that they were the loan applicants. 
These workers provided textual descriptions highlighting reasons for loan denials. These descriptions were then edited for syntactic and semantic correctness. Further, linguists also provided annotations for each description with a corresponding broad and specific reason for loan denial. For example, a broad reason could be ``job" and specific reasons could be ``no job", ``unstable job", ``limited job history", "no job history", and "unstable job history".  This resulted in a curated dataset of 2432 sentences with their corresponding broad and specific reasons. 

The number of unique reasons turned out to be less than a hundred. Thus, we observed that the space of user-friendly explanations is rather a small set. Furthermore, we observed that the reasons provided by workers seldom appeared as features in the machine learning datasets. The most frequent set of broad reasons included credit, job, income, and debt. There were a few others that were mentioned in small numbers such as failed background check, incomplete applications, etc.  

Next, in order to generate explanations serving different purposes such as those that educate and those that help in taking appropriate actions, we curated a dataset consisting of 2432 sentence pairs corresponding to two different purposes of action and education. This was done in collaboration with a linguist (Ph.D Rhetoric). We refer this dataset as the  ``{\bf Extended X-Net}''. A sample education-action explanation pair from Extended X-Net is provided below: \\ {\it The record of finances associated with this application suggests that there is a record of outstanding loan payments} (educates) \\
{\it Please complete all remaining loan payments before applying for a new loan} (suggests action). 

\section{Method} 
Our first goal is to be able to automatically generate user-friendly explanations such as in X-Net. Further, in order to ensure that explanations can be generated in a controlled manner, we want conditional generation, i.e., the rendered explanation should match with a specified reason for loan denial. Our second goal is to be able to generate explanations serving different purposes, such as those that educate and those that help in taking an action. 

Motivated by the recent success of generative adversarial networks, we design a conditional GAN in order to address the aforementioned goal. We also incorporate the GAN model with a style transfer mechanism so as to be able to generate explanations serving different purposes. However, the biggest challenge we face is the issue of limited training data (2432 sentences with <100 unique reasons). Below we describe our solution strategy.

\subsection{Network Architectures for Conditional Generation}
In this section, we describe the design choices used to improve the quality of generated textual information given limited training data. We adopt the ARAE-GAN \cite{zhao2017adversarially} architecture and build our model on top of that. ARAEGAN consists of a discrete structure encoder and a continuous space decoder that are jointly trained with a discriminator to agree in data distribution. This approach allows one to utilize a complex encoder model such as RNN and still constrain it with a very flexible but more limited generator distribution. The full model can be then used as a latent variable GAN model. In order to address the limited training data issue, we propose three modifications to  \cite{zhao2017adversarially} as listed below:

\subsubsection{Gaussian Mixture as the Random Vector}
Since curating a novel real-world text dataset is a very costly process (given the time and cost of manual annotation), it is crucial to be able to develop a model which can learn from a small size dataset.
Accordingly, to address the limited data size issue and improve the expressive power of our model, we use the key idea from \cite{gurumurthy2017deligan}, i.e. to consider a Gaussian mixture model as the noise input of the generator. Intuitively, since mixture models can approximate
arbitrarily complex latent distributions given a sufficiently large number of Gaussian components, by modeling the latent space as a mixture of learnable Gaussians instead of the conventional distribution, we can improve the expressiveness of the model. 

\subsubsection{Hierarchical Two Level Conditioning}
The idea here is to incorporate broad and specific reasons for loan denials in a hierarchical manner. The motivation comes from the fact that children learn from very limited data in a hierarchical manner \cite{kemp2007learning}. Accordingly, in addition to previous technique, to further increase the model's ability to embed the reasons and generate more relevant sentences, we first consider two level reasoning for each sentence. Then, after embedding these reasons using the encoder component of the autoencoder, we combine them into a single vector using a neural network representing their hierarchical graphical model. And finally, we incorporate the resulting vector as the conditional information to our GAN model. 

\subsubsection{Classification Losses}
In addition to the loss functions proposed in \cite{zhao2017adversarially}, to generate more relevant sentences (corresponding to conditional information), we introduce two new losses \cite{kocaoglu2017causalgan} to classify the real and generated sentences based on their reasons.  Similar to the labeler and anti-labeler loss proposed in \cite{kocaoglu2017causalgan}, we consider two classifiers on real and generated sentences and learn them simultaneously with our model. The labeler provides an estimate of the reason based on the X-Net dataset whereas the anti-labeler estimates the reason in the generated sentences. A block diagram illustrating the architecture is provided in Figure 1.
\begin{figure}[ht]
    \centering
        \includegraphics[width=\columnwidth]{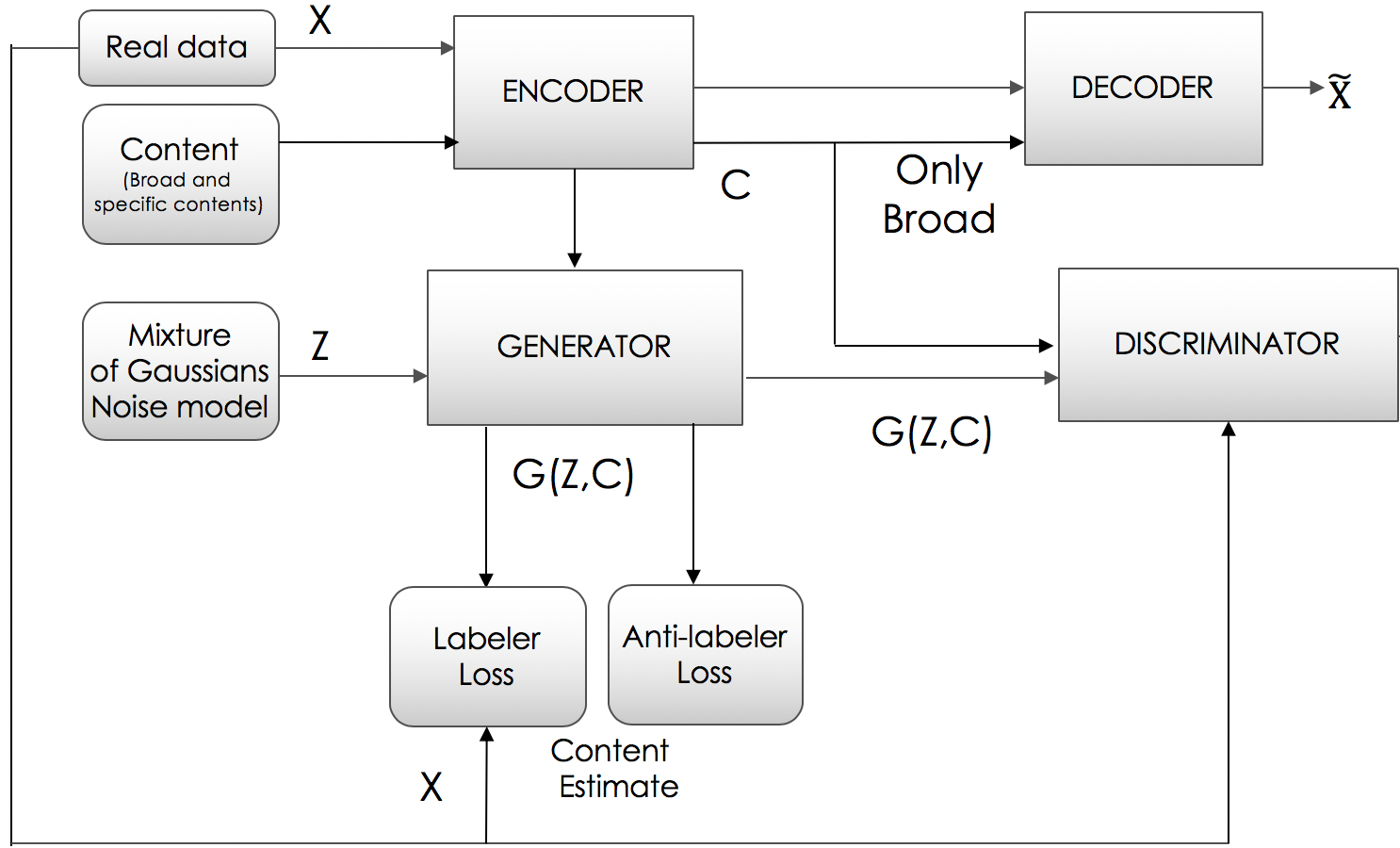}
    	\caption{Block diagram illustrating the system architecture.}
\end{figure}
\subsection{Network Architecture for Style Transfer}
The goal here is to generate sentences across different styles, namely those that educate the loan applicant about the denial and those that help them to take future actions. Limited training data poses a big challenge in this task as well. We consider three scenarios to check what best addresses the challenge. 
\subsubsection{Unaligned style transfer} First, we consider the ARAEGAN model wherein they do unaligned style transfer, i.e., without having corresponding pairs of sentences in the two styles (education and action) in the training data.  The architecture and training is the same as that proposed in \cite{zhao2017adversarially}.
\subsubsection{ARAEGAN With Gaussian mixture model for noise}
Next, we incorporate the Gaussian mixture model to the noise input of the generator in the aforementioned ARAEGAN model to check for any improvement in the result. 
\subsubsection{Aligned Style Transfer}
We consider aligned style transfer wherein we are provided with pairs of explanations corresponding to ``education'' and ``action'' in the training data. We consider two decoders one each for action and education, and a single encoder. The output of the encoder is passed to both the decoders in order to ensure alignment. The rest of the architecture and training is the same as that proposed in \cite{zhao2017adversarially}.

\section{Experiments}
In this section, we first elaborate the details of hyperparameters and model configurations. Then we evaluate the ability of our model to utilize the reason information in generating meaningful and relevant sentences using perplexity and accuracy of pre-trained classifiers as evaluation metrics. 

\subsection{Experiment Setup Details}
For a fair comparison, we implement all methods using identical loss functions and optimization for training.
We tune all the hyperparameters on the validation data and use grid search to find the best hyperparameters, such as \textit{number of gaussion mixture components}$=\{10,20,50,100\}$, $maxlen=[20,30]$, embedding dimension $d=\{100, 150, 200, 300\}$, and learning rates.
We find that the following combination of parameters works well on our dataset. Embedding size $300$, batch size $128$, maxlen $23$, and number of Gaussian mixture components $50$.
For the rest of hyperparameters, we use the same values as in \cite{zhao2017adversarially}. 

\begin{figure}[ht]
    \centering
        \includegraphics[width=\columnwidth]{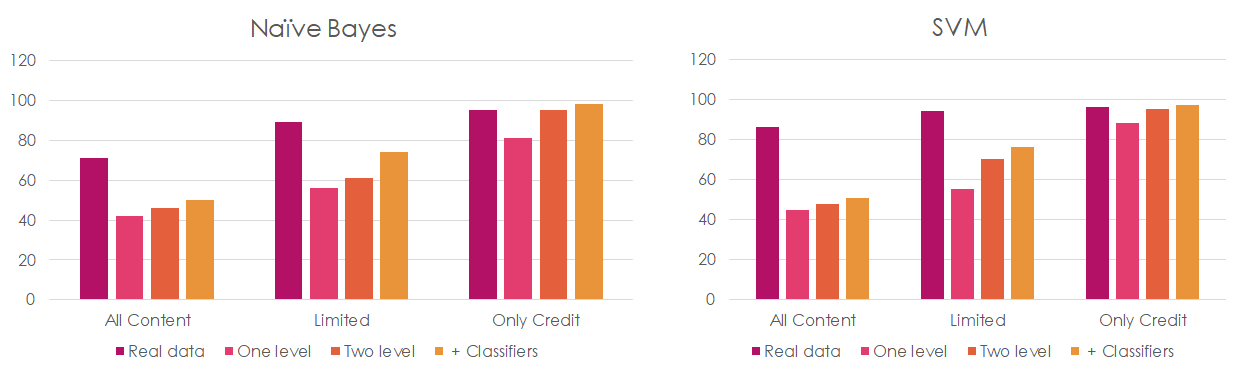}
    	\caption{The accuracy result of the SVM and Naive Bayes pre-trained classifier on different models .}
    	\label{fig:acc}
\end{figure}
\begin{table}[tbh]
	\centering
	\caption{The PPL score of different models.}
	\label{tab:PPL}
\begin{tabular}{lc}
\toprule
Models& Perplexity score \\ 
\midrule
ARAE-GAN & 10.4\\ 
ARAE-GAN+ Gaussion Mixture (GM) &7.3\\ 
ARAE-GAN+ GM + One level reason (1L) & 6.4\\ 
ARAE-GAN+ GM + Two level reasons (2L) & 6.1\\ 
ARAE-GAN+ GM + 2L + Classifiers (C) & 6.15\\ 
\bottomrule
\end{tabular}
\end{table}
\begin{figure}
    \centering
        \includegraphics[width=\columnwidth]{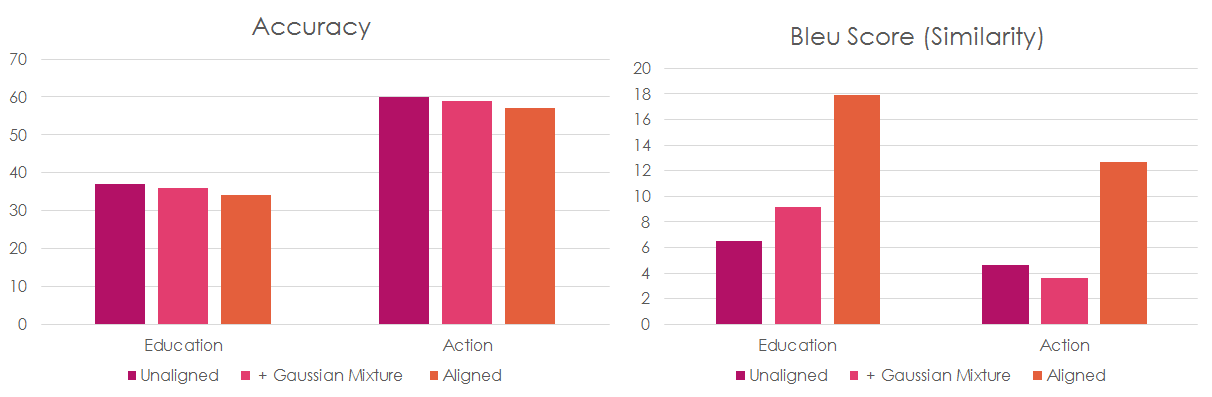}
    	\caption{The evaluation results of style transfer task.}
    	\label{fig:style}
\end{figure}
\subsection{Experimental Results}
In this section, we evaluate the capabilities of our proposed model in conditional text generation and style transfer tasks.
\subsubsection{Conditional Text Generation}
To assess the quality of generated sentences with respect to the specified reasons for loan denials, we consider three separate test data. For the first scenario, we randomly choose $100$ samples from dataset as test data and treat the rest as training data. We refer this as all content set. Furthermore, since the distribution of reasons is not uniform in the dataset, we evaluate our model by only considering 100 samples of the most appeared reasons ([`credit', `job', `debt', `income']) in our second test dataset. We call this as limited reasons set. And finally, we consider a test data containing only 'credit' (the most frequent reason) as the conditional information. The results of pre-trained SVM and Naive Bayes classifier is provided in Figure~\ref{fig:acc}. Furthermore, the perplexity score for different models is provided in Table~\ref{tab:PPL}. As shown, all models provide comparable accuracy to real data, with the final model (i.e. the model of ARAEGAN with Gaussian mixture model for noise input, coupled with two level conditioning of reasons along with labeler and anti-labeler classifiers loss functions) outperforming the rest with a considerable gap. The best result can be observed for credit-only data as it is the most frequent data instance.          


\subsubsection{Style Transfer}
Similar to the previous section, in order to evaluate the style transfer task, we randomly choose $100$ samples from dataset as test data and treat the rest as training data. The accuracy results of the classifier (which is part of the ARAE-GAN model) and bleu score is provided in Figure~\ref{fig:style}. As is shown, the aligned version of model outperforms the rest with a considerable gap.


\begin{table}[tb]
	\centering
	\caption{Generated samples for broad reason 'credit' and specific reason 'low credit'. GM denotes Gaussian mixture model for noise, 1L is one level conditioning, 2L is two level conditioning and C denotes labeler and anti-labeler classifiers.}
	\label{tab:CG-S}
\begin{tabular}{l|c}
\toprule
Models&Sentences \\ 
\midrule
ARAE-GAN &The loan is denied because the applicant is not enough the \textcolor{red}{income}.
\\ \hline
ARAEGAN+ GM &The loan is denied because the applicant's \textcolor{red}{income} is too low.
\\ \hline
ARAEGAN+ GM +1L &The loan is denied because the applicant has a poor \textcolor{ForestGreen}{credit} history.
\\ \hline
ARAEGAN+ GM + 2L &The loan is denied because the applicant's \textcolor{ForestGreen}{credit} score is \textcolor{ForestGreen}{low}.
\\ \hline
ARAEGAN+ GM +1L + C &The loan is denied because the applicant has a \textcolor{ForestGreen}{low credit} score.
\\ 
\bottomrule
\end{tabular}
\end{table}
\begin{table}[tb]
	\centering
	\caption{Style transfer samples--- Row 1 is the real sentence corresponding to training data. Row 2-4 shows the generated sentences. The bold text in row 1 denotes the reason for loan denial. The red and green colors in row 2-4 denote the (incorrect and correct) reasons in the generated sentences. GM denotes Gaussian mixture model for noise.}
	\label{tab:ST}
\begin{tabular}{l|p{5cm}|p{5cm}}
\toprule
Models&Transfer from Education to Action & Transfer from Action to Education \\ 
\midrule
Reference Sentence &there is a record of inconsistent loan \textbf{payments}. &please re-consider applying for a loan of a different \textbf{amount} that may better align with your \textbf{income}.
\\ %
\hline
Unaligned &please re-consider applying for a loan of a different amount that may better align with your \textcolor{red}{income}.&the applicant has only been \textcolor{red}{employed} at their current employer for a limited period of time.
\\ \hline

Unaligned + GM &talk to your bank about finding ways to improve your \textcolor{red}{credit}.&the \textcolor{ForestGreen}{income} associated with this application is, unfortunately, not high enough to be considered eligible for this loan.
\\ 
\hline
Aligned &maintain a consistent record of timely loan \textcolor{ForestGreen}{payments} moving forward.
&the \textcolor{ForestGreen}{income} listed on this application is not high enough to match the \textcolor{ForestGreen}{amount} requested for a loan.
\\ 
\bottomrule
\end{tabular}
\end{table}
\subsubsection{Generated samples}
The generated samples for conditional text generation is provided in Table~\ref{tab:CG-S}. As is shown, the ARAE-GAN model was incapable of generating semantically correct sentences due to the limited data issue. Furthermore, although incorporating noise as a mixture of Gaussians helped in generating more semantically correct sentences, it could not capture the broad and specific reasons (highlighted in red). As we can see, the one-level condition could only capture the broad reason but only with two-level conditioning and the two-level conditioning with classifiers model, both the broad and specific reasons could be accurately captured (highlighted in green). 

The generated samples for style transfer is provided in Table~\ref{tab:ST}. As is shown, the unaligned (ARAE-GAN) model was incapable of transferring the sentence to its corresponding style (although it could capture the style but couldn't capture the correct reasoning in the reference sentence). Along the same lines, although incorporating Gaussian mixture idea improved the semantic correctness of the generated sentences, it failed to capture the correct reason in the generated sentences (highlighted in red). On the other hand, our aligned model not only generated meaningful sentences with correct style but could also transfer the reasons correctly from the reference sentences (highlighted in green).

\section{Conclusions}
In this work, we explored the problem of explainability from an end-user perspective. In particular, we considered the use case of explaining loan denials and collected a first-of-its-kind dataset that is representative of user-friendly explanations. We observed that the reasons acceptable to users seldom appear as features in machine learning datasets and thereby contributes to the gap in understanding model-based explanations. Thus, we designed a novel conditional GAN to generate user-friendly textual explanations based on the reasons specified in the collected dataset. Additionally, the proposed model was also embedded with style transfer to generate explanations serving different purposes---to educate and to help take actions. We hope our work will help bridge the gap between research and practice in the financial industry by catering to the needs of the wider community of users seeking explanations, and by generating multiple explanations serving different purposes.
\section{Acknowledgments}
We would like to thank Cathrine Dam, Nahla Sturm-Wang and Jenae Cohn for their help in annotating the data. We also thank AMT respondents for participating in the survey.
\bibliography{reference}
\bibliographystyle{plainnat}
\end{document}